# Generating gradients in the energy landscape using rectified linear type cost functions for efficiently solving 0/1 matrix factorization in Simulated Annealing

Makiko Konoshima[1*], Hirotaka Tamura[2], and Yoshiyuki Kabashima[3]

*1 Quantum Laboratory, Fujitsu Limited, Kawasaki 211-8588, Japan*
*2 DXR Laboratory Inc., Yokohama 223-0066, Japan*
*3 Institute for Physics of Intelligence & Department of Physics, The University of Tokyo, Bunkyo, Tokyo 113-0033, Japan*

The 0/1 matrix factorization defines matrix products using logical AND and OR as product-sum operators, revealing the factors influencing various decision processes. Instances and their characteristics are arranged in rows and columns. Formulating matrix factorization as an energy minimization problem and exploring it with Simulated Annealing (SA) theoretically enables finding a minimum solution in sufficient time. However, searching for the optimal solution in practical time becomes problematic when the energy landscape has many plateaus with flat slopes. In this work, we propose a method to facilitate the solution process by applying a gradient to the energy landscape, using a rectified linear type cost function readily available in modern annealing machines. We also propose a method to quickly obtain a solution by updating the cost function's gradient during the search process. Numerical experiments were conducted, confirming the method's effectiveness with both noise-free artificial and real data.

## 1. Introduction

0/1 matrix factorization aims to find $W \in \{0,1\}^{MK}$ and $H \in \{0,1\}^{KN}$ such that $V \approx W \circ H$ using a matrix $V \in \{0,1\}^{MN}$, where the rank number $K$ in low-rank approximation of the matrix is less than $M$ or $N$. Here, the symbol $\circ$ represents the matrix product using logical operations, namely $\wedge$ for multiplication and $\vee$ summation. One of the purposes is to compress the information held by $V$ and reduce the number of elements by factorizing $V$ with $K$ smaller than $M$ and $N$, thereby efficiently performing analysis such as classification and collaborative filtering. For example, a matrix $V$ where the rows and columns represent the users and audiovisual works such as movies or DVDs, with the elements' values representing the audiovisual works' rating results. Each element's value is 0 if it is not interested or not seen if the rate is low, and 1 otherwise. The rows of $W$ are the user's feature vector, and the corresponding $H$ is the coefficient to be multiplied by $W$ to compute $V$. The matrix $W$ makes it easier to find similar users by checking the Hamming distance to determine if it is a logical value rather than a real number. If $V$ and $W$ are logical values, $H$ can be processed in the annealing machine using logic operations, which have much less computational overhead than real number operations.

In collaborative filtering, which recommends movies to users with similar viewing preferences, the Hamming distance between rows in matrix $W$ is utilized. For movies that users have not yet watched, their ratings can be estimated to predict how they might rate them. This approach is valuable for estimating movie preferences. Similarly, customer purchase behavior can be represented as a matrix that recommends products they have not purchased.

Several previous studies on matrix factorization using logical operations. Miettinen et al.[1] assumed that the rows of matrix $V$ represent instances and the columns represent features. They initially determined the presence or absence of correlations between instances as the values of elements in matrix $W$. Then, they optimized matrix $H$ to maximize the similarity between $V$ and $W \circ H$. Zhang et al.[2] performed non-negative matrix factorization with real-valued elements and subsequently adjusted the threshold for each element to determine whether it should be 0 or 1.

In such research, despite user preferences being initially determined by nonlinear decision processes such as logical operations, the computation of real values may introduce errors not originally present. An alternative approach involves using an annealing machine for matrix factorization, constraining the elements of $W$ to non-negative real numbers and those of $H$ to 0 or 1[3,4]. However, these methods require optimizing $W$ outside the annealing machine and then optimizing $H$ within the machine. This can lead to complex back-and-forth processing when using hardware or software modules for annealing, which are separated from other processes.

Nevertheless, when considering the use of a machine based on classical, i.e., non-quantum, digital circuits through a quantum-inspired annealing approach[5,6,7,8], the energy function was predominately confined to binary-quadratic forms. The reason for this is that by dealing with the binary-quadratic form, the energy difference can be easily calculated as a linear combination of variables, and efficient parallel processing dedicated to its minimization is possible.

On the other hand, schemes that extend the interaction between variables to encompass many bodies, like the extended Ising machine for Digital Annealers[9], have been proposed to improve formulation flexibility of conventional machines that deal with binary-quadratic forms. In this proposal, interactions between variables are extended to many-body interactions to enhance the computational capabilities. By applying rectified linear types and other nonlinear activation functions on the linear and logical sums of variables, which can be computed rapidly, dependent variables are generated. These dependent variables enable the handling of various many-body interactions. By introducing additional dependent-variable spins, previously challenging inequality constraint penalty functions and higher-order energy functions can be optimized rapidly, as demonstrated.

In this study, we propose a new annealing method for 0/1 matrix factorization, leveraging the ability to handle nonlinear or higher-order energy functions, including inequality constraints. However, a direct formulation leads to energy landscapes with many plateaus, rendering the search process arduous. We suggest formulating the problem using rectified linear type functions as cost functions to address this issue and propose an efficient solution method. Numerical experiments are conducted to verify the effectiveness of the proposed approach.

In Section 2 of this paper, we explain the formulation and challenges associated with solving matrix factorization using SA, the underlying principle of annealing machines. In Sect. 3, we describe the rectified linear type cost function, an anticipated feature of annealing machines, and propose a method to simplify matrix factorization using this function. In Sect. 4, we demonstrate the effectiveness of the proposed method through simple numerical experiments. In Sect. 5, we verify the effectiveness of the proposed method using real-world data. We conclude in Sect. 6 with a brief summary and mention of future work.

## 2. Formulation and the Difficulty of Obtaining the Global Optimal Solution in 0/1 Matrix Factorization

When solving the 0/1 matrix factorization problem using SA, the objective is to minimize the energy function $E$ given by Equation (1). Here, $logical(A)$ is 1 if $A$ is non-zero and 0 if $A$ is zero, and $\|A\|$ represents the absolute value of $A$.

$$E = \sum_{ij} \left\| logical\left(\sum_k W_{ik} H_{kj}\right) - V_{ij} \right\|. \quad (1)$$

If we denote the elements of $W$ and $H$ as $A$ and $B$, respectively, the relationship between logical and integer calculations is expressed as $A \wedge B = AB$ and $A \vee B = A + B - AB$. In the summation calculation within the aforementioned logical function, since $\vee$ appears $K$ times, converting it to arithmetic operations introduces a term of order $K$. Furthermore, expressing it in squared absolute values introduces a term of order $2K$. When addressing this issue with an annealing machine, situations where machines cannot handle higher-order terms are resolved by converting it into a quadratic form[10]. Whether through conversion or direct calculation, the computation of higher-order terms on annealing machines or different computing platformsrequires the storage of memory for the coupling coefficients and involves extensive memory access. However, hardware solutions, such as digital annealers, present an alternative that can mitigate these burdens by introducing auxiliary variables.

In addition to the aforementioned reasons, additional difficulties arise in solving the 0/1 matrix factorization problem using an annealing machine. One such difficulty is illustrated in Fig. 1, which shows the relationship between integer and logical values in matrix multiplication. Even if the integer values are large, the logical values are constrained to 0 or 1. This graph's shape infers that the energy landscape becomes a plateau regardless of the number of elements in $W$ and $H$ that are equal to 1, making the search difficult.

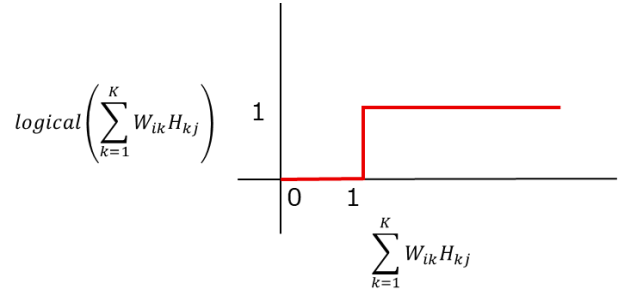

Fig. 1.  Relationship between real value and logical value in matrix multiplication.

## 3. Generation of Energy Landscape Gradient by Using Rectified Linear Type Function

We have discussed the difficulties in searching for solutions in 0/1 matrix factorization. In addition to these difficulties, solving problems involving higher-order terms is generally considered difficult due to the energy landscape exhibiting many plateaus. To address this issue, Sutton et al. proposed a method that introduces gradients to flat energy regions using Walsh transformation in MAX-k-SAT problems with higher-order terms, making it easier to obtain the global optimal solution[11]. Similarly, we believe that introducing gradients to the flat regions in a way that is consistent with the problem's nature would aid in obtaining the optimal solution. In 0/1 matrix factorization, as mentioned in Sect. 1, we consider using the rectified linear type function, which is a nonlinear function added to the annealing machine, to introduce gradients to the flat regions shown in Fig. 1. Hereafter, we refer to this function as the rectified linear type cost function or RL-type cost function. The RL-type cost function is defined such that the output is always 0 if the input value $a$ is less than or equal to 0, and it outputs $a$ directly if $a$ is greater than 0. This function is also known as the $max(0, a)$ function and is commonly used as a constraint term in constraint satisfaction problems, where constraints, such as the amount of cargo in transportation or time restrictions for movement from one point to another, ensure that the energy does not exceed a certain positive value when the constraints are not violated. The reason for using the RL-type cost function lies in its similarities with the rectified linear unit (ReLU) function, which is widely used in neural networks. This function is easy to implement and computationally efficient, benefiting from the accumulated knowledge and optimization techniques in the field.[12)]

In this work, we set the RL-type cost function and its weight $\lambda_{ij}$ for each element in accordance with the value of $\widehat{V_{ij}}$, which is the element-wise value of the matrix product in the arithmetic operation of $W$ and $H$, and the value of $V_{ij}$. The formulation is shown in Equation (2), where the RL-type cost function is described as a max function for clarity. $G_{ij}$ is the element-wise value that applies a gradient to the energy landscape using the RL-type cost function. The minimum energy remains unchanged in both (1) and (2), and if $V_{ij} = \widehat{V_{ij}}$, then $G_{ij} = 0$.

$$\widehat{V_{ij}} = \sum_{k=1}^{K} W_{ik} H_{kj}.$$

$$G_{ij}(\widehat{V_{ij}}) = \lambda_{ij} \max\left(0, (1 - V_{ij})\widehat{V_{ij}} + V_{ij}(1 - \widehat{V_{ij}})\right) = \begin{cases} \max(0, \widehat{V_{ij}}) & (V_{ij} = 0) \\ \max(0, 1 - \widehat{V_{ij}}) & (V_{ij} = 1) \end{cases}. \quad (2)$$

$$E = \sum_{ij} G_{ij}.$$

In Fig. 2, the relationship between $\widehat{V_{ij}}$ and $G_{ij}$ when performing matrix multiplication is illustrated. The larger the number of elements in $W$ and $H$ that are equal to 1 when $V_{ij} = 0$, the larger $G_{ij}$ becomes. In **Fig. 3**, a comparison of Equations (1) and (2) is made when taking the Hamming distance from the correct answer as the x-axis. Note that when using the RL-type cost function, there is a significant gradient towards the correct answer compared with when it is not used.

In Equation (2), $E$ represents the sum of the RL-type penalty functions. The number of penalty functions is equal to the number of elements in matrix $V$. When there are multiple constraints, it becomes very difficult to obtain a solution that satisfies all penalty functions if a common penalty coefficient $\lambda$ is used for all penalty functions. This is because satisfying one constraint before reaching the optimal value can lead to multiple other constraints becoming unsatisfied. Such a situation becomes a local minimum that hinders reaching the global optimum.

An effective method to escape from such local minimum is to control the penalty coefficients for each constraint. By increasing the penalty coefficient for unsatisfied constraints, at some point, the energy decrease in resolving that penalty violation surpasses the energy increase in other constraints becoming unsatisfied, and the state becomes a non-local minimum. Therefore, an algorithm was also considered where the penalty coefficients for all constraints start from the same value in the initial state, and the coefficient $\lambda$ is increased at a constant rate for unsatisfied constraints while keeping it constant for satisfied constraints.

Hereafter, the method of performing SA with Equation (1) will be referred to as BC (Binary Cost), the method of performing SA with Equation (2) without changing all $\lambda$ values uniformly will be referred to as RL-F (keep the $\lambda$ fixed), and the method of increasing $\lambda$ for unsatisfied constraints will be referred to as RL-U (update $\lambda$).

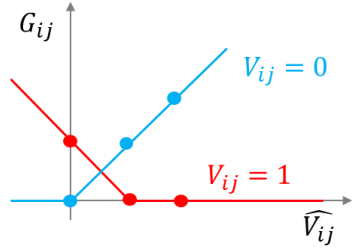

Fig. 2. Relationship between $\widehat{V_{ij}}$ and $G_{ij}$. The blue line represents $V_{ij} = 0$ with a gradient of $\lambda$. When $V_{ij} = 1$, the red line is 1 if the input value is 0, and 0 if the input value is 1 or greater.

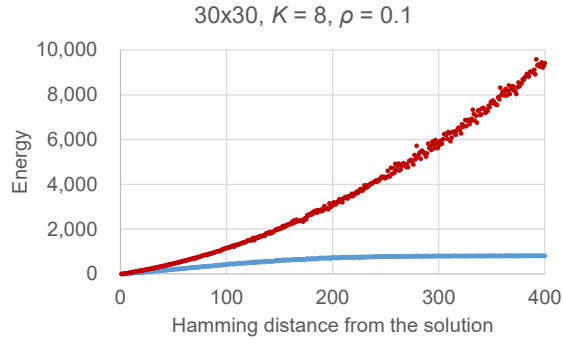

Fig. 3. Relationship between the Hamming distance from the solution and the energy is depicted. The energy for the blue dots was calculated using Eq. (1), while for the red dots, Eq. (2) (RL-F) was used. Without the use of an RL-type cost function, the energy landscape tends to flatten out into a plateau. However, by incorporating an RL-type cost function, we can observe a gradient leading towards the correct zero point.

## 4. Numerical Experiments Using Artificial Data

We investigated the number of Monte Carlo steps ($MCS0$) required to obtain a solution for BC, RL-F, and RL-U. In this work, we define a Monte Carlo step (MCS) as the number of attempts made for all spins, i.e., the elements of $W$ and $H$, once. The temperature control is performed as follows: the initial value of the inverse temperature is $\beta_0$, and the update coefficient is $\beta_f$. The temperature is updated as $\beta \leftarrow \beta(1 + \beta_f)$. The temperature is updated every 1000 acceptance counts in accordance with the Metropolis criterion, and if the rejection count reaches 1000 MCS, the process is stopped as no solution is obtained. For RL-F, $\lambda$ is the same for all $i$ and $j$. For RL-U, only the $\lambda$ values of the elements of $V$ that do not match $W \circ H$ are increased every 1 MCS, starting from an initial value $\lambda_0$, as $\lambda \leftarrow \lambda(1 + \lambda_p)$. The pseudocode for this algorithm is shown in Table I.

Table I. Energy calculation algorithm for 1 MCS. Steps 10b to 13b are applied in cases where only the element $\lambda$ of $V$, which does not match with $W \circ H$, needs to be increased.

| (a) BC | | (b) RL-F and RL-U | |
|---|---|---|---|
| 1a: | for i = 1 ... M do | 1b: | for i = 1 ... M do |
| 2a: |   for j = 1 ... N do | 2b: |   for j = 1 ... N do |
| 3a: |     randomly flip one bit of W or H | 3b: |     randomly flip one bit of W or H |
| 4a: |     $\widehat{V_{ij}} = \sum_{k=1}^{K} W_{ik} H_{kj}$ | 4b: |     $\widehat{V_{ij}} = \sum_{k=1}^{K} W_{ik} H_{kj}$ |
| 5a: |     $M_{ij}(\widehat{V_{ij}}) = logical(\widehat{V_{ij}})$ | 5b: |     $G_{ij}(\widehat{V_{ij}}) = \lambda_{ij} \max\left(0, (1 - V_{ij})\widehat{V_{ij}} + V_{ij}(1 - \widehat{V_{ij}})\right)$ |
| 6a: |   end for | 6b: |   end for |
| 7a: | end for | 7b: | end for |
| 8a: | $E = \sum_{i,j} M_{ij}$ | 8b: | $E = \sum_{i,j} G_{ij}$ |
| 9a: | decide whether to accept W and H | 9b: | decide whether to accept W and H |
| | | 10b: | if $G_{ij} \neq 0$ and if RL-U |
| | | 11b: |   $\lambda_{ij} \leftarrow \lambda_{ij}(1 + \lambda_p)$ |
| | | 12b: |   Redo the calculation of $G$ and $E$ |
| | | 13b: | end if |

For the instances, $W$ and $H$ are generated using a uniform random number $R \in \{0,1\}$ such that $V$ matches $\rho \pm 0.01$, where $\rho$ is a predetermined value. Here, $\rho$ represents the ratio of the number of elements in $V$ that are equal to 1. The instance parameters cover $(M, N) = (30, 30)$ and $(40, 40)$, and $\rho = 0.1, 0.5$, and $0.8$. For each parameter, we investigated 10 instances with 100 patterns, each with 10 initial random numbers. In particular, for $(M, N) = (40, 40)$ and $\rho = 0.1$ in BC, there were combinations of initial values and instances for which the solution could not be obtained for a long time. Therefore, 80% of the $MCS0$ values from the smallest ones were used, and for the remaining, 90% of the $MCS0$ values from the smallest ones were used. For the statistical analysis, we used the median and interquartile range as the error. We covered $\beta_0 = \{10.0, 2.0, 1.0\}$ and $\beta_f = \{0.01, 0.1\}$ to determine the values of $\beta_0$ and $\beta_f$ that result in smaller $MCS0$. The experimental results are shown in Fig. 4.

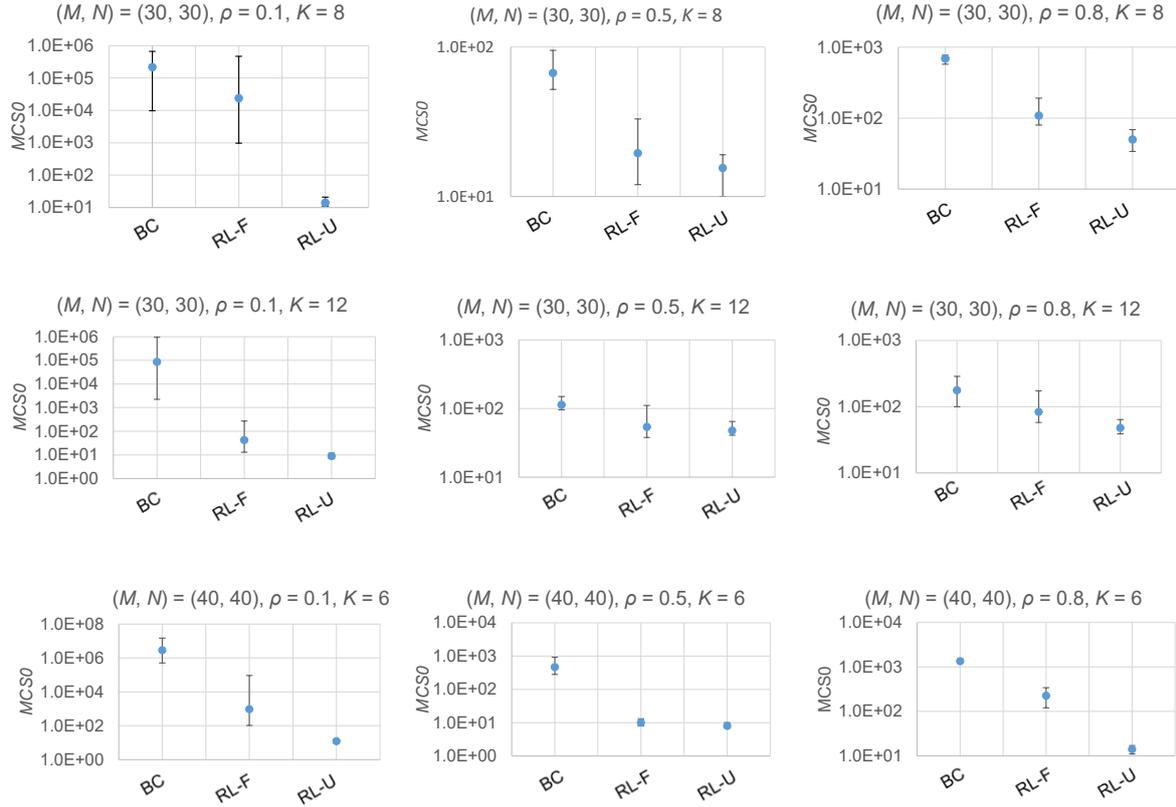

Fig. 4. ($M$, $N$) = (30, 30) (top, middle) and ($M$, $N$) = (40, 40) (bottom). On the same graph, BC and RL-F at $\lambda$ = 2.0 and RL-U at $\lambda_0$ =2.0, are arranged from left to right. The data points represent the median values, and the error range represents the quartile range.

From the experimental results, note that for $\rho$ = 0.1, RL-U has a significant effect in terms of problem solving speed in all cases. For example, $MCS0$ reaches the values in BC below 1/100 for ($M$, $N$) = (30, 30) and $K$ = 8 when $\lambda$ is increased, and below 1/1,000 for $K$ = 12. However, for smaller $MCS0$ values in BC, such as ($M$, $N$) = (30, 30), $K$ = 12, $\rho$ = 0.5 and 0.8, there is no significant difference between RL-U and other methods. Moreover, when $\lambda$ is fixed, increasing it beyond 2 does not significantly change the trends. Additionally, regardless of the experimental results, the error range of $MCS0$ decreases when $\lambda$ is increased from $\lambda_0$. Fig. 5 illustrates the relationship between the number of elements with differences in energy and $V$ and $\hat{V}$ for each step for ($M$, $N$) = (30, 30), $\rho$ = 0.1, and $K$ = 8. Note that RL-U achieves a rapid decrease in energy near the solution.

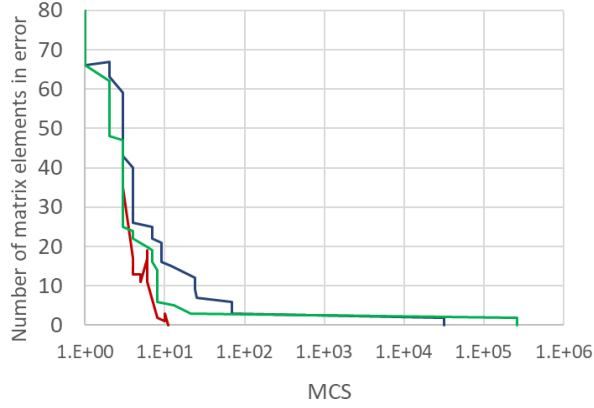

Fig. 5. Relationship between MCS and the number of matrix elements in error is depicted in the case of $(M, N) = (30, 30)$, $\rho = 0.1$, $K = 8$. BC, RL-F, and RL-U are represented by the green, blue, and red lines, respectively. Each graph is plotted using an average sample.

In these experiments, the reason that MCS0 decreases as $\rho$ increases can be explained as follows: First, for the same $V$, there are at least ${}_KC_2$ possible solutions for $W$ and $H$. This is because the matrix product remains the same when the columns of W and the rows of H are interchanged in the same order. In the ∨ calculation, all results except for 0 ∨ 0 become 1, so the number of solutions is greater than this combination count. As $\rho$ increases, the number of 1s in $W$ and $H$ also increases, leading to a larger number of solutions. The equation $V = W \circ H$ can be seen as a system of linear equations with elements being 0 or 1 and operations being logical operations. When there are more 1s, it becomes easier to represent a row of $V$ as a combination of other rows, resulting in a larger number of solutions.

Next, as a case closer to real-world problems, we conducted missing value estimation using previous artificial data. With a missing ratio of $miss = 0.1$, we randomly selected elements to be excluded from the calculation in (2). We performed a search for local solutions over 5,000,000 MCS and considered the solution with the minimum energy obtained as the estimated solution. Table II shows the error rate $error(\%)$ for the missing positions when estimating the missing values, replacing them with 0, 1, and elements randomly selected using a uniform random number with a probability $\rho$ of generating 1. Note that the error is smaller when estimating the missing values compared with replacing them, indicating the effectiveness of the estimation.

Table II. The median *error* (%) of missing values in artificial data when estimated and replaced with missing elements.

| ρ | | Estimated | Replaced | | |
|---|---|---|---|---|---|
| | | | 0 | 1 | Random |
| 0.1 | BC | 4.44 | 8.89 | 91.11 | 18.33 |
| | RL-F | 4.44 | | | |
| | RL-U | 3.33 | | | |
| 0.5 | BC | 2.78 | 52.2 | 47.8 | 53.3 |
| | RL-F | 3.33 | | | |
| | RL-U | 2.22 | | | |

## 5. Numerical Experiments Using Real Data

We conducted a numerical experiment using real data by performing 0/1 matrix factorization on the movie rating data from MovieLens (http://grouplens.org/datasets/movielens/, MovieLens 25M Dataset, released in December 2019). This dataset consists of ratings given by 162,000 users to 62,000 movies, ranging from 0.5 to 5 with increments of 0.5. The dataset is pre-tagged with genre information for each movie, such as 'Action' or 'Comedy' etc., indicating the genre of the movie. Here, we aim to identify users with similar rating patterns and determine whether to recommend movies from the unseen movies or missing data. We created $V$ from the dataset, where users are represented as rows, movies as columns, and ratings as element values. For the element values of $V$, we set ratings of 1.0 or higher as 1, and the remaining as 0. A value of 1 corresponds to having some interest or giving a non-low rating, while 0 corresponds to having no interest or giving a low rating. The 0/1 matrix factorization using AND/OR is useful for analyses such as recommendation indicating interest, because if there is at least one item of interest, the score value will remain logical 1, regardless of how many other items of interest that are logical 1 are stacked. The rows of $W$ obtained from the factorization represent features related to users' movie preferences, while $H$ represents the coefficients corresponding to those features. Users with similar Hamming distances between the row vectors of $W$ can be considered to have similar preferences. By focusing on the rows of $V$ or $\hat{V}$ for those users and summing them in the column direction, it is possible to avoid recommending movies with small values or movies with distinctive tags, and instead recommend movies with large values or other relevant movies.

In numerical experiments, processing becomes difficult with large matrix sizes. Therefore, we created a matrix using users with IDs ranging from 1 to 300 and movies with IDs ranging from 1 to 400. To exclude users and movies with fewer than 20 elements with a value of 1, we removed rows and columns with fewer than 20 elements. As a result, we obtained a final data of size $(M, N) = (74, 51)$. The $\rho$ value of this data is 0.45. In real data, factorizing $V$ into $W$ and $H$ is not always possible. Therefore, we performed a search for 10,000 $MCS$ to find the local minimum. The estimated solution obtained from this search is considered as the estimated solution, and the number of $MCS$ required to obtain this solution is referred to as $MCS1$.

We present the results of matrix factorization without missing values in Fig. 6. All three methods, BC, RL-F, and RL-U, show smaller errors compared with the previous research on discretized non-negative matrix factorization (DNMF)[2]. Note that RL-U, RL-F, and BC have fewer steps, indicating their effectiveness. Next, we introduced missing values and calculated the error for the missing positions. The results are shown in Table III. Similar to Table II, note that the estimated values have smaller errors compared with uniformly setting the missing elements to 0, 1, or randomly replacing them. This demonstrates the effectiveness of the estimation method.

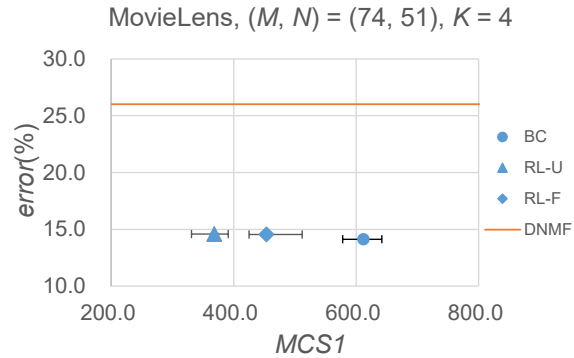

Fig. 6. (Left) Relationship between MCS1 and error in the case of the MovieLens dataset, with (M, N) = (74, 51) and k = 4. BC (circle), RL-F (rhombus), RL-U (triangle), and DNMF are shown. The DNMF is represented by an orange line, indicating only the error.

Table III. The median $error$ (%) of missing values in MovieLens dataset when estimated and replaced with missing elements.

|      | Estimated | Replaced |       |        |
|------|-----------|----------|-------|--------|
|      |           | 0        | 1     | Random |
| BC   | 32.10     | 43.50    | 56.50 | 53.32  |
| RL-F | 32.63     |          |       |        |
| RL-U | 32.63     |          |       |        |

Next, we show an example of recommendation using the RL-U results. First, the Hamming distance between user features is computed using $W$ and shown as a heat map in Fig. 7(a).

Since users with small Hamming distance have common preferences, common interest groups can be identified from the Hamming distance. For example, we can see that the user represented by column 4 forms a group with the users represented by columns 52, 54, 58, and 59 (Fig. 7((b)). We can see that the five movies that received low ratings from this common interest group (called user group #4 for short) are all in the romance genre. On the other hand, the movies in the romance genre that were watched by all members of user group #4 are also watched at high rates by members outside of this group, indicating that these movies are popular or trending movies, which are called current movies. In addition, these high-rated movies have several genres other than "romance". Therefore, we can assume that the members of user group #4 are not interested in movies that are mainly about "romance" (Table IV). Based on these results, it can be inferred that the members of this group would not recommend movies with "romance" as the main genre.

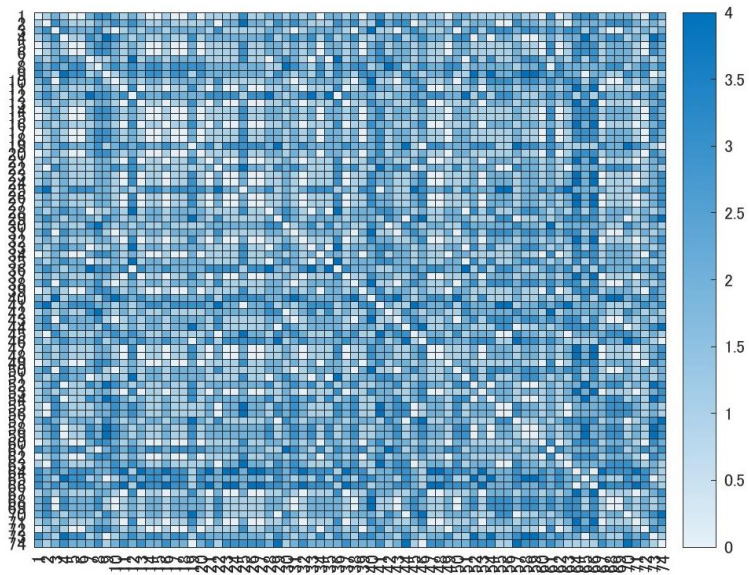

(a)

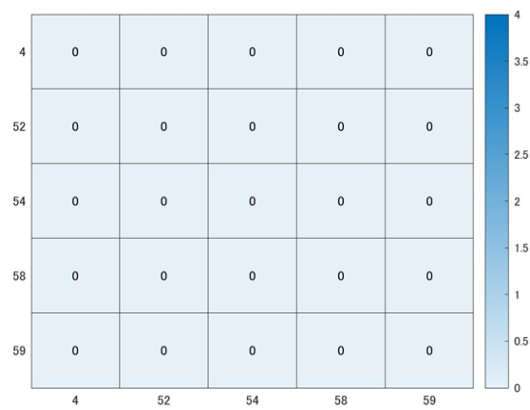

(b)

Fig. 7. Correlation between users as measured by Hamming distance: (a) among all users and (b) among users in rows 4, 52, 54, 58, and 59.

Table IV. (left) MovieIDs and genres in $V$ that were not watched or rated low by all users in group #4. (right) MovieIDs and genres in V that were watched or rated high by all users in group #4. A higher sum of $\boldsymbol{\rho}$ values in the MovieID column of $V$ indicates a higher percentage of users in $V$ who watched or rated the movie.

| Movie ID | genres |
|---|---|
| 11 | Comedy\|Drama\|Romance |
| 17 | Drama\|Romance |
| 161 | Drama\|Thriller\|War |
| 339 | Comedy\|Romance |
| 357 | Comedy\|Romance |

| Movie ID | genres | ρ of column V |
|---|---|---|
| 231 | Adventure\|Comedy | 0.41 |
| 292 | Action\|Drama\|Sci-Fi\|Thriller | 0.50 |
| 296 | Comedy\|Crime\|Drama\|Thriller | 0.68 |
| 318 | Crime\|Drama | 0.68 |
| 344 | Comedy | 0.42 |
| 356 | Comedy\|Drama\|Romance\|War | 0.64 |
| 377 | Action\|Romance\|Thriller | 0.82 |
| 380 | Action\|Adventure\|Comedy\|Romance\|Thriller | 0.59 |

In the previous analysis of the MovieLens data, a relatively large value of $\rho = 0.45$ was used. This is because the 0/1 matrix factorization using AND/OR operations has the property of recommending items if even one item can be recommended, so a large value of $\rho$ is appropriate. On the other hand, small values of $\rho$ are not expected to be suitable for 0/1 matrix factorization. Next, we provide an example where good results were not obtained for the estimation of missing values with a lower $\rho$. For the element values of $V$, we set ratings of 4.5 or higher as 1, and the remaining as 0. The resulting values are $\rho = 0.26$ and $(M, N) = (50, 42)$. The experimental results are shown in Table V. In cases where $\rho$ is small, replacing missing values with 0 did not significantly change the error. This is consistent with the results from the artificial data in Table II, where even in cases with small $\rho$, replacing missing elements with 0 resulted in several differences but not significant ones.

Table V. The median $\boldsymbol{error}$ (%) of missing values in MovieLens dataset, in the case of $\boldsymbol{\rho}$=0.26, when estimated and replaced with missing elements.

| | Estimated | Replaced | | |
|---|---|---|---|---|
| | | 0 | 1 | random |
| BC | 28.10 | 24.29 | 75.71 | 39.52 |
| RL-F | 24.76 | | | |
| RL-U | 24.76 | | | |

## 6. Conclusion

We proposed a formulation using RL-type cost functions to facilitate the solution of problems that result in energy landscapes that are difficult to solve using SA, such as 0/1 matrix factorization. In numerical experiments, we observed that using the proposed method resulted in a reduced number of MCS required to find the solution. RL-type cost functions are expected to be widely adopted in the future for solving real-world problems in Ising machines, and there are high expectations for their widespread adoption and utilization in various devices in the future. Note that it is also possible to consider modifying the function applied to the approach of altering the energy landscape to make it easier to solve without changing the minimum energy. This presents a potential avenue for future research.

*E-mail: makiko@fujitsu.com